\documentclass[10pt,twocolumn,letterpaper]{article}

\usepackage{iccv}
\usepackage{times}

\usepackage{graphicx,amsmath,amssymb,booktabs,tabulary,multirow,array,float,algorithm,algorithmic}
\usepackage[accsupp]{axessibility}

\newcolumntype{I}{!{\vrule width 3pt}}
\newlength\savedwidth

\newlength\savewidth
\newcommand\shline{\noalign{\global\savewidth\arrayrulewidth
\global\arrayrulewidth 1pt}%
\hline
\noalign{\global\arrayrulewidth\savewidth}}

\newcolumntype{x}[1]{>{\centering\arraybackslash}p{#1pt}}

\newcommand{\tablestyle}[2]{\setlength{\tabcolsep}{#1}\renewcommand{\arraystretch}{#2}\centering\footnotesize}

\usepackage[pagebackref=true,breaklinks=true,letterpaper=true,colorlinks,bookmarks=false]{hyperref}

\iccvfinalcopy 

\def\iccvPaperID{5626} 

\ificcvfinal\fi

\begin{document}

\title{Pseudo-label Alignment for Semi-supervised Instance Segmentation}

\author{
Jie Hu$^{1\dagger}$, Chen Chen$^{1\dagger}$, Liujuan Cao$^1$\thanks{Corresponding author. $^{\dagger}$Equal contribution.}, Shengchuan Zhang$^1$, Annan Shu$^2$,\\ Guannan Jiang$^2$, and Rongrong Ji$^1$\\
$^1$Key Laboratory of Multimedia Trusted Perception and Efficient Computing, \\Ministry of Education of China, Xiamen University\\
$^2$Contemporary Amperex Technology Co. Limited
}

\maketitle
\ificcvfinal\thispagestyle{empty}\fi

\begin{abstract}
Pseudo-labeling is significant for semi-supervised instance segmentation, which generates instance masks and classes from unannotated images for subsequent training.
However, in existing pipelines, pseudo-labels that contain valuable information may be directly filtered out due to mismatches in class and mask quality.
To address this issue, we propose a novel framework, called pseudo-label aligning instance segmentation (PAIS), in this paper.
In PAIS, we devise a dynamic aligning loss (DALoss) that adjusts the weights of semi-supervised loss terms with varying class and mask score pairs.
Through extensive experiments conducted on the COCO and Cityscapes datasets, we demonstrate that PAIS is a promising framework for semi-supervised instance segmentation, particularly in cases where labeled data is severely limited.
Notably, with just 1\% labeled data, PAIS achieves 21.2 mAP (based on Mask-RCNN) and 19.9 mAP (based on K-Net) on the COCO dataset, outperforming the current state-of-the-art model, \ie, NoisyBoundary with 7.7 mAP, by a margin of over 12 points.
Code is available at: \url{https://github.com/hujiecpp/PAIS}.
\end{abstract}

\section{Introduction}
Semi-supervised instance segmentation aims to alleviate the significant burden of human labeling by utilizing a small amount of labeled data in conjunction with abundant unlabeled data~\cite{a8578531,wang2022noisy,xu2021end}.
Existing semi-supervised instance segmentation pipelines typically generate pseudo-labels from unlabeled images, which are then used to train the models together with labeled images.
Therefore, pseudo-labels play a crucial role in semi-supervised instance segmentation.
The generation of pseudo-masks, pseudo-classes, and pseudo-boxes from unlabeled images improves the model training.
However, current semi-supervised instance segmentation frameworks do not fully leverage the potential of such pseudo-labels.
Specifically, pseudo-labels with mismatched class and mask scores are often filtered out by fixed thresholds, leading to the exclusion of valuable information that could aid the model training.
For instance, pseudo-labels with high-quality masks but low class scores would be filtered out by a class threshold, resulting in the loss of the pixel-level information.

\begin{figure*}
\centering
\includegraphics[width=1.0\linewidth]{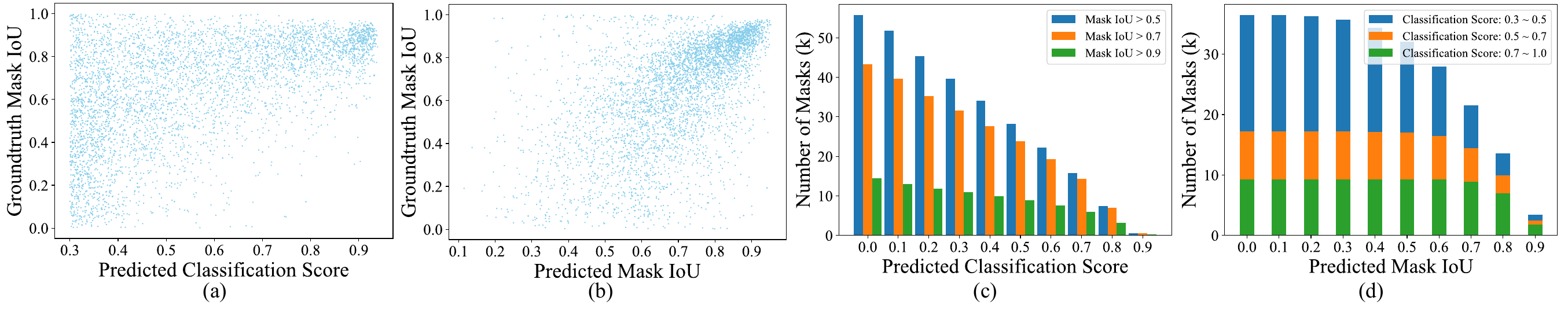}\vspace{-2mm}
\caption{
\textbf{Pseudo-label wasting in semi-supervised instance segmentation.}
To illustrate this issue, we employed the K-Net model~\cite{zhang2021k} trained on 10\% labeled images and randomly sampled 5k unlabeled images, from the COCO \texttt{train2017} dataset, to plot the figures.
In (a), we present the predicted classification score \wrt the ground truth mask intersection over union (IoU), revealing that the predicted classification score is inadequate in evaluating mask quality.
In (b), we added a mask IoU prediction branch to the model and plotted the predicted mask IoU \wrt the ground truth mask IoU, demonstrating that the predicted mask IoU can represent the mask quality.
However, as shown in (c), a significant number of pseudo-labels with high-quality masks, \ie, mask IoU $>$ 0.7, will be filtered out with high classification thresholds.
Additionally, (d) shows that pseudo-labels with high-quality masks can also have correspondingly low classification scores.
Therefore, the misalignment between the mask and class scores leads to a significant number of valuable pseudo-labels being excluded during semi-supervised training.
}\label{fig1}
\end{figure*}
In this paper, we present a new semi-supervised framework, termed pseudo-label aligning for instance segmentation (PAIS), aiming to improve the utilization of filtered pseudo-labels.
As illustrated in Fig.~\ref{fig1}, the main challenge of PAIS lies in the mismatched scores between pseudo-classes and pseudo-masks.
The classification score and mask intersection over union (IoU) are misaligned in assessing the quality of pseudo-labels.
As a result, masks with high IoUs would be filtered out due to low classification scores, and vice versa.
In the mean time, lowering the threshold of both scores introduces incorrect classes or low-quality masks into the semi-supervised training.
To overcome this dilemma, we propose a dynamic aligning loss (DALoss) that softly re-weights the classification and the segmentation losses based on the quality of different pseudo-labels.
Specifically, DALoss penalizes low-score pseudo-labels rather than filtering them and promotes high-score ones, to adjust their contribution to the final loss function.
Our experiments on the COCO and Cityscape datasets demonstrate the effectiveness of the proposed PAIS framework.
Specifically, with only 1\% labeled data, PAIS achieves 21.2 mAP and 19.9 mAP on the COCO dataset using Mask-RCNN~\cite{he2017mask} and K-Net~\cite{zhang2021k}, respectively.
This outperforms the current state-of-the-art model, NoisyBoundary~\cite{wang2022noisy}, by more than 12 points.

Our contributions can be summarized as follows:
\begin{itemize}
\vspace{-2mm}
\item We propose a novel pseudo-label aligning framework for semi-supervised instance segmentation, called PAIS, which unleashes the potential of utilizing pixel-level pseudo-labels in semi-supervised instance segmentation.
Furthermore, to the best of our knowledge, PAIS is the first framework that can be adapt to box-free instance segmentation models.
\vspace{-2mm}
\item We introduce a new loss function, named dynamic aligning loss (DALoss), which incorporates pseudo-labels with diverse class and mask qualities into the semi-supervised training process.
DALoss consistently enhances the performance of box-free and box-dependent instance segmentation frameworks.
\vspace{-2mm}
\item We conduct comprehensive experiments on the COCO and Cityscapes datasets to evaluate PAIS.
In particular, PAIS achieves state-of-the-art results on the COCO dataset, \ie, 19.9, 27.6, and 31.1 mAP for the box-free pipeline K-Net~\cite{zhang2021k}, and 21.2, 29.3, and 31.1 mAP for the box-dependent pipeline Mask-RCNN~\cite{he2017mask}, with 1\%, 5\%, and 10\% labeled data, respectively.
\end{itemize}
\begin{figure*}
\centering
\includegraphics[width=1.0\linewidth]{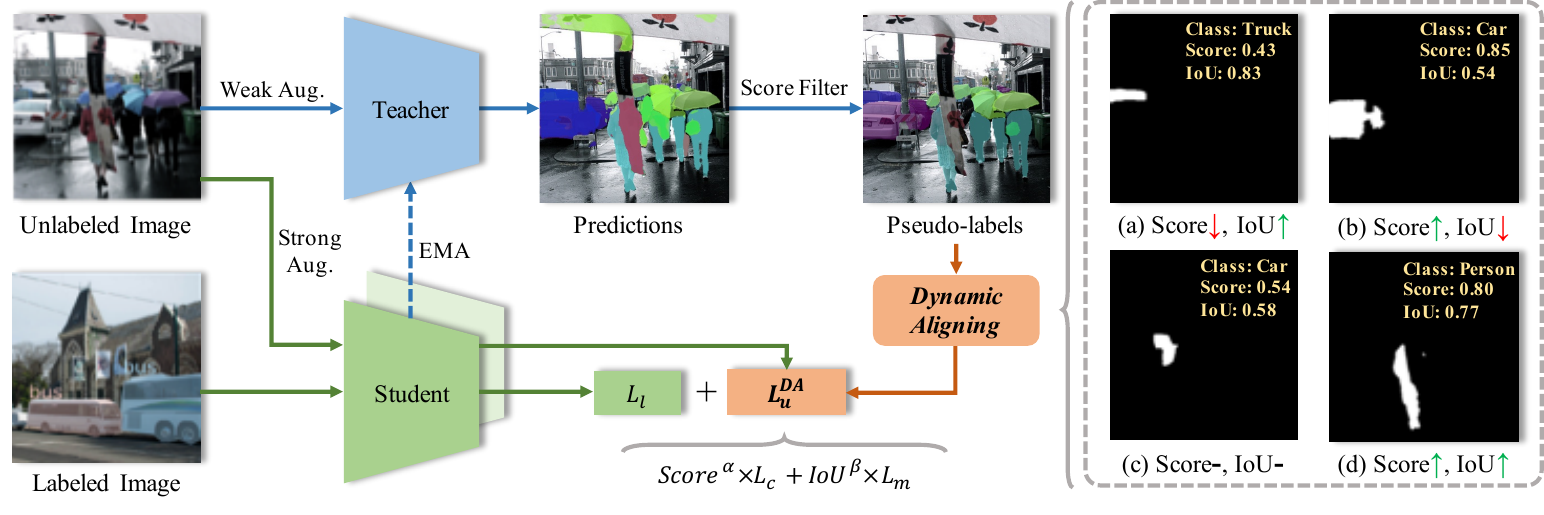}\vspace{-2mm}
\caption{
\textbf{Framework of pseudo-label aligning for semi-supervised instance segmentation (PAIS).}
The focus of PAIS is to explore the utility of pixel-level pseudo-labels for instance segmentation.
To this end, we propose a novel approach called dynamic aligning loss (DALoss), which enables the incorporation of pseudo-labels with varying quality during training.
This is particularly significant when the number of labeled images is limited. 
With DALoss, a significant number of predictions from the teacher model can be leveraged in the semi-supervised training process, rather than being discarded.
Specifically, we consider three types of pseudo-masks with varying quality: (a) pseudo-masks with incorrect classes but high-quality masks, \ie, low classification scores but high mask IoUs; (b) pseudo-masks with accurate classes but low-quality masks, \ie, high classification scores but low mask IoUs; and (c) pseudo-masks with medium qualities.
%
%
}\label{fig2}
\end{figure*}
\section{Related Work}
\textbf{Semi-supervised Image Classification.} 
In image classification, semi-supervised learning has been extensively explored, and the methods can be classified into two categories: pseudo-label-based and consistency-regularization-based methods.
Specifically, pseudo-label-based methods~\cite{scudder1965probability,lee2013pseudo} leverage pre-trained models to generate annotations for the unlabeled images to train the model.
In contrast, consistency-regularization-based methods~\cite{bachman2014learning,rasmus2015semi,laine2016temporal,sajjadi2016regularization,xie2020unsupervised,berthelot2019remixmatch,dai2017good,ma2023xmesh,ma2022xclip,li2019attribute} incorporate various data augmentation techniques such as random regularization~\cite{french2017self} and adversarial perturbation~\cite{miyato2018virtual} to generate different inputs for one image and enforce consistency between these inputs during training.
FixMatch~\cite{sohn2020fixmatch} combines the consistency-regularization-based techniques with a pseudo-label-based framework by applying a strong-weak data augmentation pipeline to input images and enforcing consistency between the augmented images. In this work, we follow the pseudo-label-based methods and also use strong-weak data augmentation during training in PAIS.

\textbf{Semi-supervised Object Detection.}
From the very beginning, STAC~\cite{sohn2020simple} proposed the use of pseudo-labels and consistency training for semi-supervised object detection.
However, the effectiveness of the method was limited by the two-stage training pipeline similar to that of Noisy Student~\cite{xie2020self}, where the pseudo-labels were generated from pre-trained models and were not updated along with the model training.
After STAC, several studies~\cite{xu2021end,zhou2021instant,tang2021humble,yang2021interactive,liu2021unbiased} incorporated the idea of exponential moving average (EMA) from MeanTeacher~\cite{tarvainen2017mean}.
The teacher model and pseudo-labels are updated after each training iteration to generate instant pseudo-labels, making the entire pipeline end-to-end trainable.
Additionally, Unbiased Teacher~\cite{liu2021unbiased} utilized Focal loss~\cite{lin2017focal} instead of traditional cross-entropy loss to alleviate the problem of unbalanced pseudo-labels.
In this paper, we also follow the idea of incorporating EMA into the proposed PAIS framework, with a focus on integrating pixel-level annotations into the training process.

\textbf{Fully-supervised Instance Segmentation.}
Instance segmentation aims to provide pixel-level predictions for each object instance in an image.
Existing methods can be classified into three categories: top-down (or box-dependent) methods, bottom-up methods, and direct segmentation (or box-free) methods.
Top-down methods~\cite{hu2021ISTR,liu2023grounding,ren2023strong,ren2023detrex, lu2023candy} such as Mask R-CNN~\cite{he2017mask}, YOLACT~\cite{bolya2019yolact}, and CenterMask~\cite{lee2020centermask} generate bounding boxes first and then segment the objects within the boxes.
Bottom-up methods~\cite{yuan2020deep,de2017semantic,liu2017sgn,gao2019ssap,hu2023you}, regard instance segmentation as a label-then-cluster problem, which classify each pixel first and then group the pixels into an arbitrary number of object instances.
Direct segmentation or box-free methods such as SOLO~\cite{wang2020solo,wang2020solov2}, K-Net~\cite{zhang2021k}, MaskFormer~\cite{cheng2021per,cheng2021masked} and SOTR~\cite{guo2021sotr} deal with instance segmentation without bounding box detection.
Any of the aforementioned instance segmentation methods can be implemented into PAIS, and in this work, we present two examples, one from the box-dependent category, Mask R-CNN, and another from the box-free category, K-Net.

\textbf{Semi-supervised Instance Segmentation.}
Semi-supervised instance segmentation is commonly considered to be a sub-task of semi-supervised object detection~\cite{xu2021end, liu2021unbiased}.
Consequently, existing frameworks rely heavily on bounding boxes, in which segmentation performance is strongly dependent on detection performance.
Among them, NoisyBoundary~\cite{wang2022noisy} was the first to formally propose the semi-supervised instance segmentation task.
Recent efforts have been made to construct box-free pipelines for fully-supervised instance segmentation~\cite{wang2020solo,wang2020solov2,zhang2021k,cheng2021per}.
In this paper, we investigate PAIS in both box-free and box-dependent instance segmentation, revealing the potential of fully utilizing pixel-level annotations.
In contrast to the recently-proposed PoliteTeacher~\cite{filipiak2022polite}, which filters out pseudo-labels with low confidence, the proposed PAIS leverages them. It is a novel and effective way of utilizing noisy pseudo-labels for semi-supervised learning.

\section{Method}
\subsection{Task Formulation}
The goal of PAIS is to better leverage unlabeled images with a limited number of pixel-labeled images to boost the performance of semi-supervised instance segmentation.
The PAIS framework consists of three key steps: (1) pseudo-label generation, (2) dynamic pseudo-label alignment, and (3) end-to-end model training.
In the pseudo-label generation step, we introduce a mask scoring branch~\cite{huang2019mask} that predicts mask IoUs as an additional metric along with classification scores to assess the quality of pseudo-labels.
In the dynamic aligning step, we re-weight the loss terms based on the quality of different pseudo-labels.
Finally, the teacher and student models are trained using exponential moving average (EMA)~\cite{tarvainen2017mean}.
The overall framework of PAIS is illustrated in Fig.~\ref{fig2}, and we introduce the key steps in detail as follows.
It is important to note that the PAIS framework can be applied to any box-dependent or box-free framework for enhanced exploitation of pseudo-labels in semi-supervised instance segmentation.
We instantiate two examples by Mask-RCNN~\cite{he2017mask} and K-Net~\cite{zhang2021k} in this paper, but do not restrict to them.
\subsection{Pseudo Label Generation}
In the pseudo-label generation, we apply weak data augmentation to the unlabeled images and input them into a teacher model.
The weak data augmentation includes scaling, horizontal flipping, and other augmentation operations that do not alter the image's content.
The teacher model produces a set of pseudo-labels, including masks, boxes, and classification scores for each input image.
In box-free pipelines, the box predictions are optional.

As depicted in Fig.~\ref{fig1}(a)(c), using the classification score alone is inadequate to measure the quality of predicted masks.
To address this, we incorporate a mask scoring branch into the pipeline to predict mask Intersection over Union (IoU) for evaluating mask quality.
As shown in Fig.~\ref{fig1}(b), the predicted mask IoUs can be used to measure mask quality effectively, which has also been verified in MS-RCNN~\cite{huang2019mask}.
After generating pseudo-labels in the previous step, the predictions are filtered using two thresholds: a classification threshold $\tau_{cls}$ and a mask IoU threshold $\tau_{iou}$.
The resulting set of pseudo-labels includes $N$ elements, each containing a mask $\boldsymbol{m}_i\in\mathbb{R}^{H\times W}$, a bounding box $\boldsymbol{b}_i\in\mathbb{R}^{4}$, a classification score $\boldsymbol{c}_i\in\mathbb{R}^{L}$ (with an additional background category), and a mask IoU score ${s}_i\in\mathbb{R}$.
Note that $H, W$ denote the mask resolution, and $L$ is the number of classes.

\subsection{Dynamic Aligning Loss}
Although good pseudo-labels can be obtained by setting high thresholds for classification scores and mask IoUs, a large number of predictions with misaligned mask and classification qualities are discarded (as illustrated in Fig.\ref{fig1}(c)(d)).
These misaligned predictions can be useful since the amount of labeled data is limited in the semi-supervised setting.
Fig.\ref{fig2} shows examples of pseudo-labels that can be included in the training process.
If the misaligned pseudo-labels are directly used in semi-supervised learning, the incorrect predictions on classes or masks can introduce significant noise.
To reduce the noise, we propose a dynamic aligning loss (DALoss).

\textbf{Vanilla Loss for PAIS.} In semi-supervised instance segmentation, the loss function can be decomposed into two terms for labeled and unlabeled images, as:
\begin{equation}
\begin{split}
\label{eq1}
\mathcal{L}_{semi}=\lambda_l\mathcal{L}_{l}+\lambda_u\mathcal{L}_{u},
\end{split}
\end{equation}
where $\lambda_l$ and $\lambda_u$ are the hyper-parameters for balancing the loss terms.
\begin{algorithm}[t]
\caption{Pseudo-label Alignment}
\label{alg1}
\begin{algorithmic}
\STATE \textit{1:} Initialize teacher and student models randomly.
\STATE \textit{2:} \textbf{Repeat}
\STATE \textit{3:}\ \ \ \ \ Apply weak augmentation to unlabeled images.
\STATE \textit{4:}\ \ \ \ \ Obtain pseudo-labels $\{\boldsymbol{m}_i,\boldsymbol{b}_i, \boldsymbol{c}_i,{s}_i|i=1,\ldots,N\}$ from teacher model via thresholds $\tau_{cls}, \tau_{iou}$.
\STATE \textit{5:}\ \ \ \ \ Apply strong augmentation to unlabeled images.
\STATE \textit{6:}\ \ \ \ \ Obtain predictions $\widetilde{\boldsymbol{m}}, \widetilde{\boldsymbol{b}},\widetilde{\boldsymbol{c}}$ for unlabeled images.
\STATE \textit{7:}\ \ \ \ \ \ Calculate $\mathcal{L}_{u}$ for unlabeled images by Eq.~\ref{eq5}.
\STATE \textit{8:}\ \ \ \ \ \ Inference labeled images by student model.
\STATE \textit{9:}\ \ \ \ \ \ Calculate $\mathcal{L}_{l}$ for labeled images.
\STATE \textit{10:}\ \ \ \ Train student model with the loss $\mathcal{L}_{semi}$ in Eq.~\ref{eq1}.
\STATE \textit{11:}\ \ \ \ Update teacher model via EMA.
\STATE \textit{12:} \textbf{Until} scheduled epochs.
\end{algorithmic}
\end{algorithm}
The loss for labeled images can be defined using the functions commonly employed in instance segmentation, augmented with an additional binary cross-entropy term for regressing mask IoUs.
Given the pseudo-labels from the teacher model and the predictions from the student model, we formulate the loss for unlabeled images as:
\begin{equation}
\begin{split}
\label{eq2}
\mathcal{L}_{u}=\frac{1}{N_{pos}}\sum_{i=1}^{N_{pos}}\bigl(\lambda_{b}&\mathcal{L}_{b}(\widetilde{\boldsymbol{b}}_{\nu(i)}, \boldsymbol{b}_{\mu(i)}) \\
+ \lambda_{c}\mathcal{L}_{c}(\widetilde{\boldsymbol{c}}_{\nu(i)}, \boldsymbol{c}_{\mu(i)}^{o}&)
+\lambda_{m}\mathcal{L}_{m}(\widetilde{\boldsymbol{m}}_{\nu(i)}, \boldsymbol{m}_{\mu(i)}^{b})\bigr) \\
&+ \frac{1}{N_{neg}}\sum_{i=1}^{N_{neg}}\bigl(\lambda_{c}\mathcal{L}_{c}(\widetilde{\boldsymbol{c}}_{i},\boldsymbol{c}_{neg}^{o})\bigr),
\end{split}
\end{equation}
where $\mathcal{L}_{b}$ includes the box IoU loss and the L1 loss, $\mathcal{L}_{c}$ denotes the cross-entropy loss, $\mathcal{L}_{m}$ is the dice loss.
In Eq.~\ref{eq2}, $\widetilde{\boldsymbol{c}}_{i}, \widetilde{\boldsymbol{b}}_{i}, \widetilde{\boldsymbol{m}}_{i}$ represent predictions from the student model.
The pseudo-scores $\boldsymbol{c}_{\mu(i)}$ for classification are converted to one-hot vectors $\boldsymbol{c}_{\mu(i)}^{o}$ for the category with the highest score.
The pseudo-masks $\boldsymbol{m}_{\mu(i)}$ are activated by sigmoid function and discretized into binary value as $\boldsymbol{m}_{\mu(i)}^{b}$.
The one-hot vector for the background class is denoted by $\boldsymbol{c}_{neg}^{o}$.
The label indexing functions, $\nu(i)$ and $\mu(i)$, match predictions with pseudo-labels for training, which are defined in terms of different pseudo-label assigning strategies.
In our implementations, the one-to-many assignment defines the label indexing functions $\nu(i)=i$ and $\mu(i)$ as:
\begin{equation}
\begin{split}
\label{eq3}
\mu(i)=\mathop{\arg\min}\limits_{t\in[1,N]}\bigl(\mathcal{L}_{b}(\widetilde{\boldsymbol{b}}_{i}, \boldsymbol{b}_{t})\bigr).
\end{split}
\end{equation}
%
Instead, the one-to-one assignment defines the label indicting functions $\mu(i)=i$, and finds the optimal $\nu^{*}(\cdot)$ via:
\begin{equation}
\begin{split}
\label{eq4}
\nu^{*}=\mathop{\arg\min}\limits_{\nu}&\sum_{i=1}^{t}\big(\lambda_{b}\mathcal{L}_{b}(\widetilde{\boldsymbol{b}}_{\nu(i)}, \boldsymbol{b}_{i}) +\lambda_{c}\mathcal{L}_{c}(\widetilde{\boldsymbol{c}}_{\nu(i)}, \boldsymbol{c}_{i}^{o}) \\
&+\lambda_{m}\mathcal{L}_{m}(\widetilde{\boldsymbol{m}}_{\nu(i)}, \boldsymbol{m}_{i}^{b})-s_i\bigr).
\end{split}
\end{equation}
Note that the loss terms for regressing boxes are optional in box-free instance segmentation frameworks.

\textbf{Dynamic Aligning Loss for PAIS.} 
To optimize the model using pixel-level pseudo-labels, we propose to replace Eq.~\ref{eq2} with DALoss.
Since the teacher model provides ideal metrics for measuring the quality of classification scores and mask IoUs, we propose to adjust the weight of loss terms based on the quality of the pseudo-labels. This is achieved by using the following equation:
\begin{equation}
\begin{split}
\label{eq5}
\mathcal{L}_{u}^{DA}=\frac{1}{N_{pos}}\sum_{i=1}^{N_{pos}}&\bigl(\lambda_{b}\mathcal{L}_{b}(\widetilde{\boldsymbol{b}}_{\nu(i)}, \boldsymbol{b}_{\mu(i)}) \\
&+\lambda_{c}(c_{\mu(i)}^h)^{\alpha}\mathcal{L}_{c}(\widetilde{\boldsymbol{c}}_{\nu(i)}, \boldsymbol{c}_{\mu(i)}^{o}) \\
&+\lambda_{m}(s_{\mu(i)})^{\beta}\mathcal{L}_{m}(\widetilde{\boldsymbol{m}}_{\nu(i)}, \boldsymbol{m}_{\mu(i)}^{b})\bigr) \\
&+ \frac{1}{N_{neg}}\sum_{i=1}^{N_{neg}}\bigl(\lambda_{c}\mathcal{L}_{c}(\widetilde{\boldsymbol{c}}_{i},\boldsymbol{c}_{neg}^{o})\bigr),
\end{split}
\end{equation}
where $c_{\mu(i)}^{h}$ denotes the highest classification score, and $s_{\mu(i)}$ denotes the mask IoU from the $\mu(i)$-th pseudo-label, $\alpha,\beta$ are the hyper-parameters.
Specifically, Eq.~\ref{eq5} adjusts the weights for the pseudo-labels conditioned on their qualities, \ie, dynamically dependent on the input images.
For instance, for a pseudo-label with a low classification score and high mask IoU, DALoss encourages the segmentation loss while constraining the classification loss.

\subsection{End-to-End Model Training}
Inspired by~\cite{tarvainen2017mean}, we employ EMA with the strong-weak data augmentation for PAIS.
Specifically, unlabeled images undergo both strong and weak data augmentations and are then fed into the student and teacher models, respectively.
The student model is trained to produce consistent results with the pseudo-labels, and the teacher model is updated by EMA.
The training pipeline for PAIS is presented in Alg.~\ref{alg1}.
\begin{table}[t]
\centering
\tablestyle{6.6pt}{1.13}\begin{tabular}{l|c|c|c|c}\shline
Method & 1\%  & 5\%  & 10\% & 100\% \\ \hline
Mask-RCNN~\cite{he2017mask}, supervised$^*$ & 3.5 & 17.3 & 22.0 & 34.5 \\
Mask-RCNN$^\dagger$~\cite{he2017mask}, supervised$^*$ & 3.5 & 17.4 & 21.9 & 37.1 \\
DD~\cite{a8578531} & 3.8 & 20.4 & 24.2 & 35.7 \\
Noisy Boundaries~\cite{wang2022noisy} & 7.7 & 24.9 & 29.2 & 38.6 \\ \hline
PAIS, on Mask-RCNN, \textit{ours} & \textbf{21.2} & \textbf{29.3} & \textbf{31.1} &  \textbf{39.5} \\ \shline
\end{tabular} \vspace{1.5mm}
\caption{\textbf{Comparison to state-of-the-art} semi-supervised instance segmentation methods on the COCO \texttt{val2017}. $^\dagger$ denotes using the same data augmentation as semi-supervised training. $^*$ denotes data from NoisyBoundary~\cite{wang2022noisy}.}\label{tab1}
\end{table}

\begin{table}[t]
\centering
\tablestyle{7.0pt}{1.13}\begin{tabular}{l|c|c|c|c}\shline
Method & 5\%  & 10\%  & 20\% & 30\% \\ \hline
Mask-RCNN~\cite{he2017mask}, supervised$^*$ & 11.8 & 16.8 & 22.3 & 26.3 \\
Mask-RCNN$^\dagger$~\cite{he2017mask}, supervised$^*$ & 11.3 & 16.4 & 22.6 & 26.6 \\
DD~\cite{a8578531} & 13.7 & 19.2 & 24.6 & 27.4 \\
STAC~\cite{sohn2020simple} & 11.9 & 18.2 & 22.9 & 29.0 \\
CSD~\cite{JisooJeong2019ConsistencybasedSL} & 14.1 & 17.9 & 24.6 & 27.5 \\
CCT~\cite{YassineOuali2020SemiSupervisedSS} & 15.2 & 18.6 & 24.7 & 26.5 \\
Dual-branch~\cite{WenfengLuo2020SemisupervisedSS} & 13.9 & 18.9 & 24.0 & 28.9 \\
Ubteacher~\cite{liu2021unbiased} & 16.0 & 20.0 & 27.1 & 28.0 \\
Noisy Boundaries~\cite{wang2022noisy} & 17.1 & 22.1 & 29.0 & 32.4 \\ \hline
PAIS, on Mask-RCNN, \textit{ours} & \textbf{18.0} & \textbf{22.9} & \textbf{29.2} & \textbf{32.8} \\ \shline
\end{tabular} \vspace{1.5mm}
\caption{\textbf{Comparison to state-of-the-art} semi-supervised instance segmentation methods on the Cityscapes validation set. $^\dagger$ denotes using the same data augmentation as semi-supervised training. $^*$ denotes data from NoisyBoundary~\cite{wang2022noisy}.}\label{tab2}
\end{table}

\section{Experiments}
\subsection{Datasets and Evaluation Metrics}
We conducted extensive experiments on the COCO~\cite{lin2014microsoft} and Cityscapes~\cite{Cordts2016Cityscapes} datasets to study the proposed PAIS.
The COCO dataset consists of 118k images with 80-class instance labels, as well as 123k unlabeled images.
The Cityscapes dataset contains urban street-view scenes and has 8 instance categories in 2.9k training images and 0.5k validation images.
For the COCO dataset, we randomly sampled 1\%, 5\%, and 10\% of the images from the \texttt{train2017} split as labeled data and treated the rest as unlabeled data following common settings.
In addition, we also used the full COCO \texttt{train2017} as labeled data and incorporated the 123k unlabeled data from COCO \texttt{unlabel2017} to train the PAIS models.
For the Cityscapes dataset, we randomly sampled 5\%, 10\%, 20\%, and 30\% of the images from the training set as labeled data and treated the remaining as unlabeled data following the common settings.
We evaluated the PAIS models on the validation sets of the COCO and Cityscapes datasets, and reported standard COCO metrics including AP, AP${50}$, AP${75}$ (averaged over IoU thresholds), and AP$_S$, AP$_M$, AP$_L$ (AP for instances of different scales).
\begin{table*}[t]
\tablestyle{2.0pt}{1.13}\begin{tabular}{lccccccccc}\shline
\multicolumn{1}{c|}{Method} & \multicolumn{3}{c|}{1\% COCO} & \multicolumn{3}{c|}{5\% COCO} & \multicolumn{3}{c}{10\% COCO} \\ \hline
\multicolumn{1}{c|}{} & AP & AP$_{0.5}$ & \multicolumn{1}{c|}{AP$_{0.75}$} & AP & AP$_{0.5}$ & \multicolumn{1}{c|}{AP$_{0.75}$} & AP & AP$_{0.5}$ & AP$_{0.75}$ \\ \hline
\multicolumn{10}{c}{\textit{Box-free Instance Segmentation}} \\ \hline
\multicolumn{1}{l|}{K-Net~\cite{zhang2021k}, supervised} & 8.03$\pm$0.25 & 16.33$\pm$0.31 & \multicolumn{1}{c|}{7.00$\pm$0.25} & 17.4$\pm$0.22 & 30.08$\pm$0.28 & \multicolumn{1}{c|}{16.83$\pm$0.21} & 21.63$\pm$0.21 & 37.43$\pm$0.38 & 21.8$\pm$0.20 \\
\multicolumn{1}{l|}{K-Net$^\dagger$~\cite{zhang2021k}, supervised} & 11.63$\pm$0.05 & 22.30$\pm$0.08 & \multicolumn{1}{c|}{10.95$\pm$0.10} & 22.28$\pm$0.05 & 38.54$\pm$0.09 & \multicolumn{1}{c|}{22.7$\pm$0.07} & 26.53$\pm$0.12 & 44.87$\pm$0.15 & 27.20$\pm$0.17 \\ \hline
\multicolumn{1}{l|}{K-Net, PAIS w/o DALoss} & 17.77$\pm$0.06 & 32.17$\pm$0.15 & \multicolumn{1}{c|}{17.53$\pm$0.12} & 25.40$\pm$0.08 & 43.12$\pm$0.05 & \multicolumn{1}{c|}{26.05$\pm$0.13} & 29.30$\pm$0.07 & 48.64$\pm$0.05 & 30.53$\pm$0.13 \\
\multicolumn{1}{l|}{K-Net, PAIS} & \textbf{19.78}$\pm$0.10 & \textbf{35.48}$\pm$0.15 & \multicolumn{1}{c|}{\textbf{19.65}$\pm$0.06} & \textbf{27.53}$\pm$0.06 & \textbf{45.63}$\pm$0.06 & \multicolumn{1}{c|}{\textbf{28.70}$\pm$0.10} & \textbf{31.04}$\pm$0.06 & \textbf{50.50}$\pm$0.07 & \textbf{32.34}$\pm$0.05 \\ \hline
\multicolumn{10}{c}{\textit{Box-dependent Instance Segmentation}} \\ \hline
\multicolumn{1}{l|}{Mask-RCNN~\cite{he2017mask}, supervised$^*$} & 3.5 & - & \multicolumn{1}{c|}{-} & 17.3 & - &  \multicolumn{1}{c|}{-} & 22.0 & - & - \\
\multicolumn{1}{l|}{Mask-RCNN$^\dagger$~\cite{he2017mask}, supervised} & 11.54$\pm$0.09 & 19.86$\pm$0.11 &  \multicolumn{1}{c|}{11.64$\pm$0.09} & 22.35$\pm$0.06 & 37.98$\pm$0.10 & \multicolumn{1}{c|}{23.14$\pm$0.11} & 27.07$\pm$0.06 & 45.10$\pm$0.10 & 28.67$\pm$0.06 \\ \hline
\multicolumn{1}{l|}{Mask-RCNN, PAIS w/o DALoss} & 20.13$\pm$0.06 & 33.23$\pm$0.15 & 
\multicolumn{1}{c|}{21.27$\pm$0.06} & 27.36$\pm$0.06 & 44.10$\pm$0.10 & 
\multicolumn{1}{c|}{29.27$\pm$0.06} & 29.77$\pm$0.06 & 47.70$\pm$0.10 & 31.97$\pm$0.06 \\ 
%
\multicolumn{1}{l|}{Mask-RCNN, PAIS} & \textbf{21.12}$\pm$0.05 & \textbf{36.03}$\pm$0.05 & \multicolumn{1}{c|}{\textbf{22.75}$\pm$0.10} & \textbf{29.28}$\pm$0.13 & \textbf{47.25}$\pm$0.13 & \multicolumn{1}{c|}{\textbf{31.20}$\pm$0.22} & \textbf{31.03}$\pm$0.06 & \textbf{49.83}$\pm$0.12 & \textbf{33.23}$\pm$0.06 \\ \shline
\end{tabular} \vspace{1.5mm}
\caption{\textbf{Results of PAIS with extremely limited number of labeled images.}
We randomly sampled 1\%, 5\%, and 10\% of the labeled images from the COCO \texttt{train2017} dataset, and repeated each experiment three times.
The average values with standard deviation are reported.
$^\dagger$ denotes using the same data augmentation as semi-supervised training.
$^*$ denotes data from NoisyBoundary~\cite{wang2022noisy}.
We can see that PAIS achieves good performance on both box-free and box-independent instance segmentation frameworks.
}\label{tab3}
\end{table*}

\begin{table}[t]
\centering
\tablestyle{1.4pt}{1.13}\begin{tabular}{l|ccc|ccc} \shline
Method & AP & AP$_{0.5}$ & AP$_{0.75}$ & AP$_{S}$ & AP$_{M}$ &  AP$_{L}$ \\ \hline
\multicolumn{7}{c}{\textit{Box-free Instance Segmentation}} \\ \hline
K-Net~\cite{zhang2021k}, supervised &  37.8 & 60.3 & 39.9 & 16.9 & 41.2 & 57.5 \\
K-Net$^\dagger$~\cite{zhang2021k}, supervised &  38.4 & 61.4 & 40.3 & 17.6 & 41.8 & 58.0 \\ \hline
K-Net, PAIS w/o DALoss &  39.4 & 62.2 & 41.6 & 18.5 & 42.8 & 59.2  \\ \hline
K-Net, PAIS  &  \textbf{40.8} & \textbf{63.5} & \textbf{43.3} & \textbf{19.2} & \textbf{44.4} & \textbf{61.4}  \\ \hline
\multicolumn{7}{c}{\textit{Box-dependent Instance Segmentation}} \\ \hline
Mask-RCNN~\cite{he2017mask}, supervised & 37.1 & 58.5 & 39.7 & 18.7 & 39.6 & 53.9 \\
Mask-RCNN$^\dagger$~\cite{he2017mask}, supervised & 37.5 & 58.9 & 40.4 & 18.6 & 40.2 & 53.8 \\ \hline
Mask-RCNN, PAIS w/o DALoss & 38.4 & 59.7 & 41.5 & 19.4 & 41.1 & 55.0 \\ \hline
Mask-RCNN, PAIS & \textbf{39.5} & \textbf{60.6} & \textbf{43.0} & \textbf{19.9} & \textbf{42.4} & \textbf{56.6}  \\ \shline
\end{tabular} \vspace{1.5mm}
\caption{\textbf{Results of PAIS with abundant labeled images.}
We utilize all the labeled images in the COCO \texttt{train2017} dataset, and supplement them with the unlabeled images in the COCO \texttt{unlabel2017} dataset for semi-supervised training.
$^\dagger$ denotes using the same data augmentation as semi-supervised training.}\label{tab4}
\end{table}

\subsection{Implementation Details}
We provide two examples of implementing PAIS with K-Net~\cite{zhang2021k} and Mask-RCNN~\cite{he2017mask}.
The models are trained using AdamW with a learning rate of 0.0001 for K-Net, and SGD with a learning rate of 0.01 for Mask-RCNN.
The hyper-parameters $\lambda_l$ and $\lambda_u$, which balance the loss terms for labeled and unlabeled images, are set to 1.0 and 0.3 for K-Net, and 1.0 and 1.5 for Mask-RCNN.
We set the thresholds $\tau_{cls}$ and $\tau_{iou}$ experimentally as 0.35 and 0.30, respectively.
For bipartite matching loss, we use the same hyperparameters as in~\cite{zhang2021k}.
The loss balancing parameters for box, class, and mask are set as $\lambda_b$=2.0, $\lambda_c$=4.0, and $\lambda_m$=1.0, respectively.
We train the models on 4 GPUs with 4 images per GPU (1 labeled and 3 unlabeled images) for 220k iterations, unless otherwise specified.
The teacher model is updated via EMA with a momentum of 0.999.
We use ResNet50~\cite{he2016deep} as the backbone for these models.

\subsection{Main Results}
\textbf{Comparison to state-of-the-art semi-supervised instance segmentation frameworks.}
In Tab.~\ref{tab1}, we compare the performance of models trained with PAIS to the state-of-the-art semi-supervised instance segmentation frameworks on the COCO dataset.
The results demonstrate that both K-Net and Mask-RCNN trained with PAIS surpass the previous methods DD~\cite{a8578531} and Noisy Boundaries~\cite{wang2022noisy} by a large margin, especially when the number of labeled data is very limited (with only 1\% or 5\% labeled images).
Specifically, when using 1\% labeled COCO images, PAIS with K-Net achieves 19.9 mask mAP, which is 12.2 points higher than Noisy Boundaries.
Interestingly, the proposed PAIS brings about better performance for Mask-RCNN when the percentage of labeled data is 1\% or 5\%, even though it originally has a inferior performance in fully-supervised instance segmentation compared to K-Net.
To explain, the bounding boxes may provide better optimization for Mask-RCNN models when the labeled data is limited.
Finally, when using 10\% and 100\% labeled images, PAIS with K-Net outperforms PAIS with Mask-RCNN.
In Tab.~\ref{tab2}, we compare PAIS with state-of-the-art methods on the Cityscape dataset, in which PAIS also achieves better performance than the predominant models.
Specifically, we report NoisyBoundaries~\cite{wang2022noisy} w/o FocalLoss in Tab.~\ref{tab2}, as we do not apply FocalLoss in PAIS.
This ensures a fair and consistent comparison. Furthermore, we also add FocalLoss to PAIS on 10\% Cityscapes, which achieves 25.1\%, surpassing 23.7\% of NoisyBoundaries w/ FocalLoss.

The more significant performance improvement on the COCO dataset validates the effectiveness of our method for solving the noisy pseudo-label problem. The COCO dataset has 80 instance categories, while the Cityspaces dataset only has 8 instance categories. This implies that the COCO dataset can provide more diverse and informative pseudo-labels, which matches our goal to utilize noisy pseudo-labels for semi-supervised learning.

\textbf{Results with an extremely limited number of labeled images.}
To demonstrate the effectiveness of PAIS, we conduct experiments with extremely limited numbers of labeled images, as shown in Table~\ref{tab3}.
Specifically, we compare the performance of various models trained with randomly sampled 1\%, 5\%, and 10\% labeled images.
First, we train supervised models, K-Net (supervised) and Mask-RCNN (supervised), with the limited labeled images.
Second, we train the same supervised models with the same data augmentation used in the semi-supervised setting, denoted as K-Net$^\dagger$ (supervised) and Mask-RCNN$^\dagger$ (supervised).
Third, we train PAIS models without DALoss, denoted as PAIS w/o DALoss on K-Net and Mask-RCNN.
Lastly, we train the PAIS models with DALoss, denoted as PAIS on K-Net and Mask-RCNN.
All models are trained three times, and the reported results are averaged.

Based on the results of K-Net (supervised) and Mask-RCNN (supervised), it can be observed that fully-supervised models perform poorly when the number of labeled images is limited.
The results of K-Net$^\dagger$ (supervised) and Mask-RCNN$^\dagger$ (supervised) show slight improvement with weak data augmentation from the semi-supervised setting.
Interestingly, while K-Net outperforms Mask-RCNN in the fully-supervised setting, their performance is similar when the number of labeled images is limited, as indicated in the table. 
Comparing the performance of K-Net (PAIS w/o DALoss) and Mask-RCNN (PAIS w/o DALoss) with that of the supervised setting reveals significant improvement.
By introducing the dynamically re-weighting process via DALoss, the performance is further improved.
For instance, with only 1\% labeled images, the mAP of K-Net and Mask-RCNN improved from 11.63 and 11.54 to 19.78 and 21.12, respectively, resulting in approximately +8.15 and +9.58 points improvement to the performance.
Additionally, the improvement over AP$_{0.5}$ and AP$_{0.75}$ suggests that DALoss considers masks with moderate quality during training, which further benefits semi-supervised learning.

\textbf{Results with abundant labeled images.}
In Tab.~\ref{tab4}, we investigate the performance of semi-supervised learning when abundant labeled data is available.
Specifically, we use the entire COCO \texttt{train2017} dataset as labeled data and COCO \texttt{unlabel2017} dataset as unlabeled data to train the models.
The results show that the semi-supervised learning approach also leads to a performance gain.
For instance, K-Net (PAIS) achieves a performance gain of approximately 1.0 point on mAP from semi-supervised learning.
Additionally, the proposed DALoss consistently improves the performance of both box-dependent and box-free instance segmentation frameworks.

\subsection{Ablation Study}
In our ablation study, we investigate several aspects that can impact the performance of PAIS, including the utilization of various loss terms, the setting of hyper-parameters, the threshold values, the varying ratios of labeled and unlabeled images, and the convergence times.
We perform the ablation studies on the COCO dataset using K-Net and Mask-RCNN, which are trained via PAIS under the setting of 10\% labeled images.
\begin{table}[t]
\centering
\tablestyle{4.3pt}{1.13}\begin{tabular}{c|c|c|ccc|ccc}\shline
Cls. & IoU. & Mask. & AP & AP$_{0.5}$ & AP$_{0.75}$ & AP$_{S}$ & AP$_{M}$ &  AP$_{L}$ \\ \hline
&   &  & 29.3   & 48.7 & 30.5 & 11.3 & 31.4 & 45.5         \\ \hline
\checkmark &  & & 29.7 & 49.0 & 31.0 & 11.4 & 31.8 & 46.3   \\
 & \checkmark &  & 29.4  &  48.9 & 30.6 & 11.7 & 31.5 & 46.0 \\
  & \checkmark & \checkmark & 30.4 & 50.1 & 31.9 & 11.9 & 32.8 & 47.6 \\ \hline
\checkmark         & \checkmark & \checkmark & \textbf{31.1}  & \textbf{50.6} & \textbf{32.4} & \textbf{12.3} & \textbf{33.3} & \textbf{48.3}      \\ \shline
\end{tabular} \vspace{1.5mm}
\caption{ \textbf{Effectiveness of different loss terms in DALoss}. \textit{Cls.} denotes the terms of dynamic aligned classification scores. \textit{IoU.} denotes adding the mask IoU branch to the model. \textit{Mask.} denotes using the terms of dynamic aligned mask scores. }\label{tab5}
\end{table}

\begin{table}[t]
\centering
\tablestyle{7.8pt}{1.13}\begin{tabular}{c|ccc|ccc}\shline
$\alpha,\beta$ & AP & AP$_{0.5}$ & AP$_{0.75}$ & AP$_{S}$ & AP$_{M}$ &  AP$_{L}$ \\ \hline
1 & 29.9  & 49.1  &  31.0 & 10.9 & 32.1  &  47.0   \\
2 & 30.2  & 49.5  &  31.1 & 11.1 & 32.2  &  47.1   \\
3 &  30.5  & 50.1  &    31.7 & 11.6 & 32.9  &  47.6   \\
4 & \textbf{31.1} &  \textbf{50.6} & \textbf{32.4} & \textbf{12.3} & \textbf{33.3} & \textbf{48.3}   \\ \shline
\end{tabular} \vspace{1.5mm}
\caption{ \textbf{Hyper-parameters in DALoss.}
Considering that the classification and segmentation are both important, we simply set $\alpha=\beta$ to investigate the influence of hyper-parameters.
}\label{tab6}
\end{table}

\textbf{Effectiveness of different terms in DALoss.}
Tab.~\ref{tab5} shows the efficacy of the different components in DALoss, which suggests that:
(1) DALoss yields an improvement of 1.8 points to the model.
Specifically, the mAP increases from 45.5 to 48.3 on AP$_L$, indicating a significant improvement for large objects.
(2) The aligning weights for classification loss alone can also provide a marginal performance gain for the model.
(3) Simply adding a mask IoU branch to the model does not enhance the overall performance.
However, when incorporating the aligning weights for the segmentation loss, the mAP is increased.
(4) Interestingly, the mask IoU branch can help improve the AP for objects of different scales.
This is due to the mask IoU branch helping to select good pseudo-labels for training.

\begin{figure}
\centering
\includegraphics[width=0.94\linewidth]{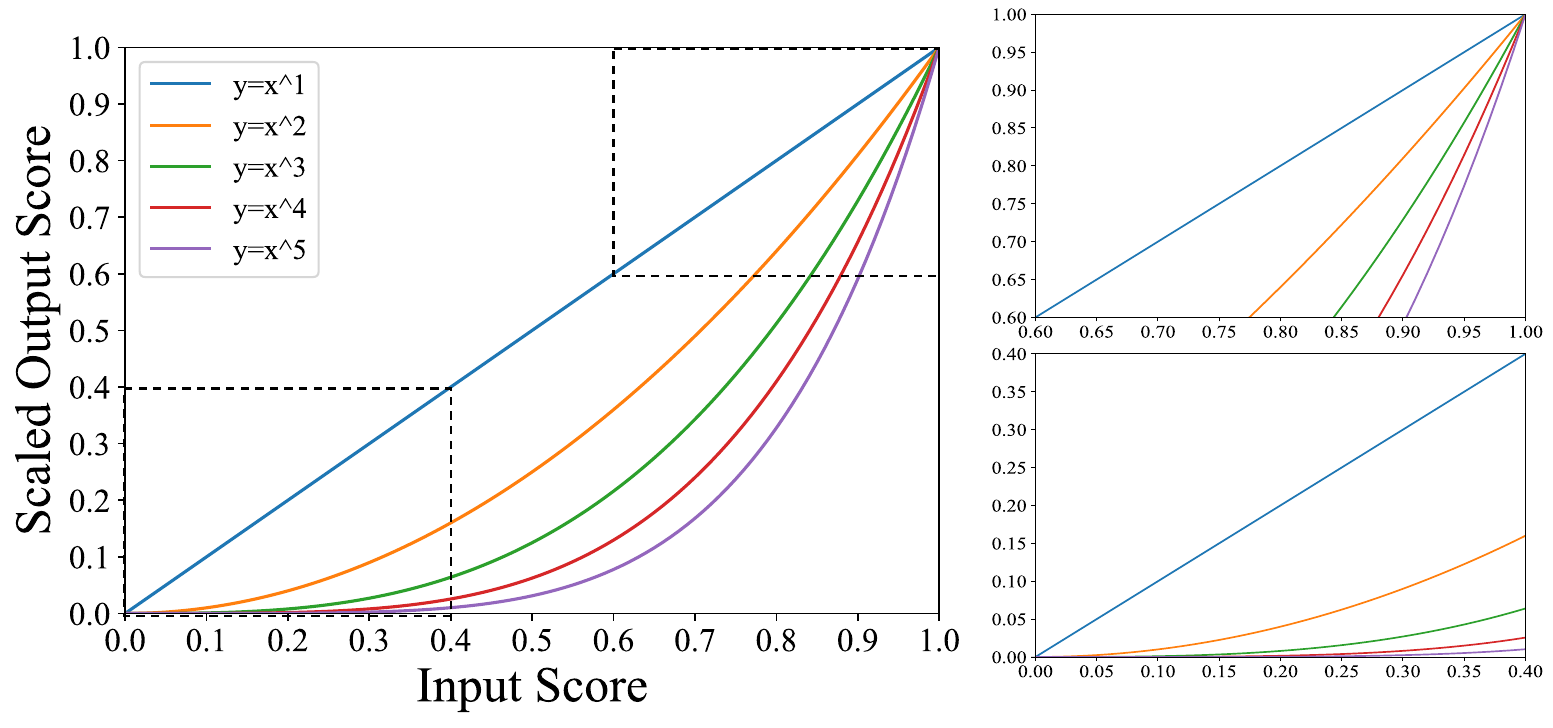}\vspace{-1mm}
\caption{
\textbf{Visualization of the loss alignment with different hyper-parameters.} When changing the hyper-parameter from 1 to 5, the input classification score or mask IoU will be more properly adjusted by their quality, \ie, the difference between high-\&low-quality mask or class will be enlarged.
}\label{fig3}
\end{figure}

\begin{figure}
\centering
\includegraphics[width=1.0\linewidth]{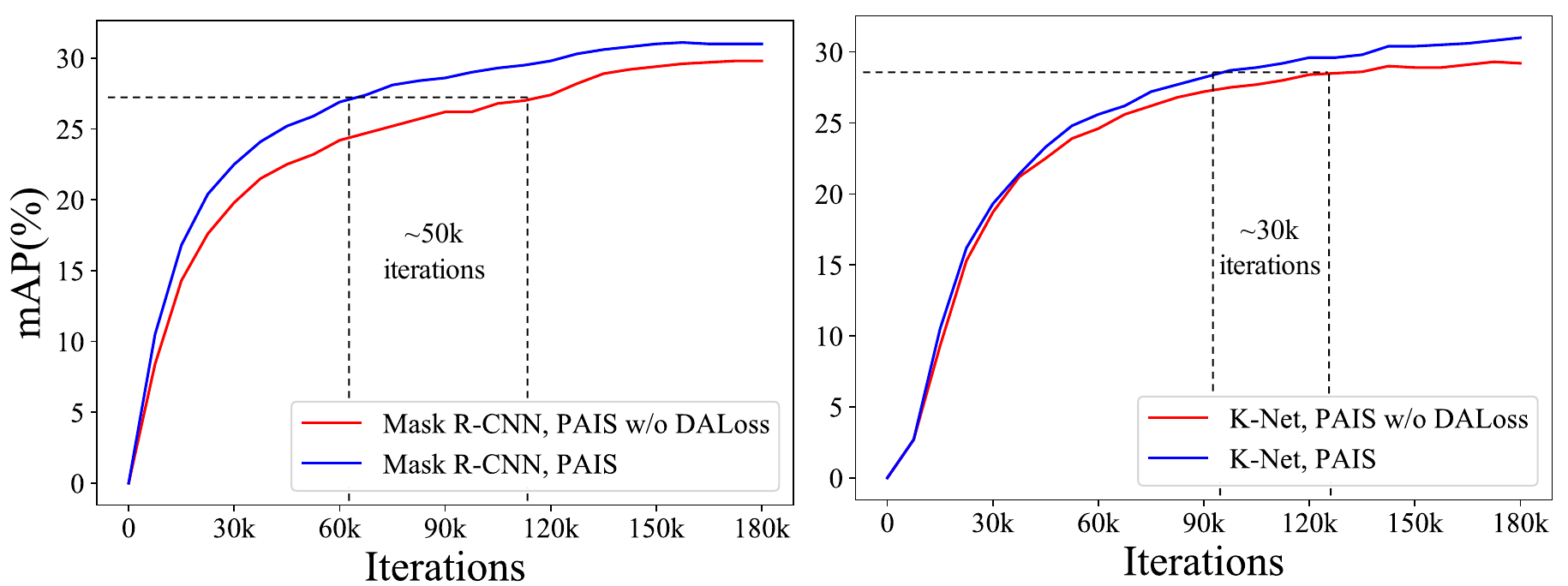}\vspace{-0.6mm}
\caption{
\textbf{Model convergence speed.} DALoss leads to faster model convergence for training.
}\label{fig4}
\end{figure}

\begin{table}[t]
\centering
\tablestyle{7.8pt}{1.13}\begin{tabular}{c|ccc|ccc}\shline
$\tau_{cls}$ & AP & AP$_{0.5}$ & AP$_{0.75}$ & AP$_{S}$ & AP$_{M}$ &  AP$_{L}$ \\ \hline
0.35 & 31.1   &  50.6   &  32.4 &  12.3 &   33.3  &   48.3 \\
0.50 &   30.6   &  50.0  &   31.9    &   11.2 & 32.6 & 47.8     \\
0.65 &  29.9  &  49.1 & 31.3 & 10.1 & 32.1 & 47.4  \\ \shline
\end{tabular} \vspace{1.5mm}
\caption{ \textbf{Results of different classification score threshold $\tau_{cls}$.} The mask IoU threshold $\tau_{iou}$ is set to 0.3.}\label{tab7}
\end{table}

\begin{table}[t]
\centering
\tablestyle{7.8pt}{1.13}\begin{tabular}{c|ccc|ccc}\shline
$\tau_{iou}$ & AP & AP$_{0.5}$ & AP$_{0.75}$ & AP$_{S}$ & AP$_{M}$ &  AP$_{L}$ \\ \hline
0.3 &  31.1   &  50.6   &  32.4 &  12.3 &   33.3  &   48.3     \\
0.5 &  30.8   &  50.3   & 31.9 & 12.0 & 32.9 & 47.9 \\
0.7 &  30.6  &  50.0 & 31.2 & 11.6 & 32.3 & 47.1  \\ \shline
\end{tabular} \vspace{1.5mm}
\caption{ \textbf{Results of different mask IoU threshold $\tau_{iou}$.}
The classification score threshold $\tau_{cls}$ is set to 0.35.
}\label{tab8}
\end{table}

\begin{table}[t]
\centering
\tablestyle{7.6pt}{1.13}\begin{tabular}{c|ccc|ccc}\shline
Ratio & AP & AP$_{0.5}$ & AP$_{0.75}$ & AP$_{S}$ & AP$_{M}$ &  AP$_{L}$ \\ \hline
1:1 & 29.0  &  47.7  &  31.0   &   11.0 & 31.6  &  45.5    \\
1:2 & 30.1  &  48.8  &  32.2   &   11.5 & 32.4  &  47.1    \\
1:3 & 31.1  &  50.6  &  32.4   &   12.3 & 33.3  &  48.3   \\
1:4 & 31.3  &  50.9  &  32.6   &   12.4 & 33.5  &  48.7      \\ \shline
\end{tabular} \vspace{1.5mm}
\caption{ \textbf{Different ratio of labeled images and unlabeled images in a batch.} 
The performance saturates when the ratio of labeled and unlabeled images is set to 1:4.
}\label{tab9}
\end{table}
\begin{figure*}
\centering
\includegraphics[width=1.0\linewidth]{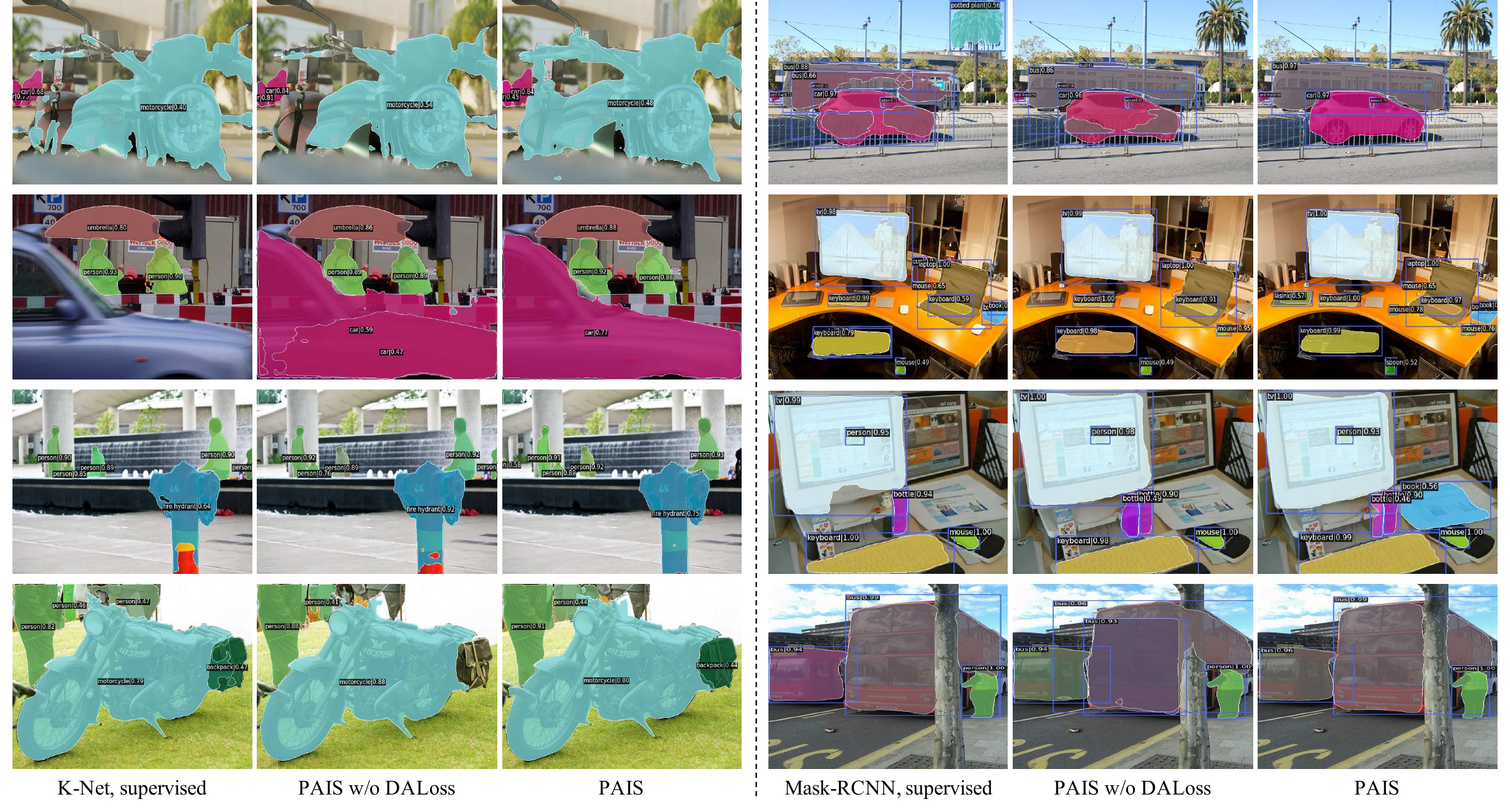}\vspace{-1.0mm}
\caption{
\textbf{Visualization of the predictions from different models.} From the examples, we can clearly show the benefits introduced by pixel-level information from PAIS and DALoss.
}\label{fig5}
\end{figure*}

\begin{figure*}
\centering
\includegraphics[width=1.0\linewidth]{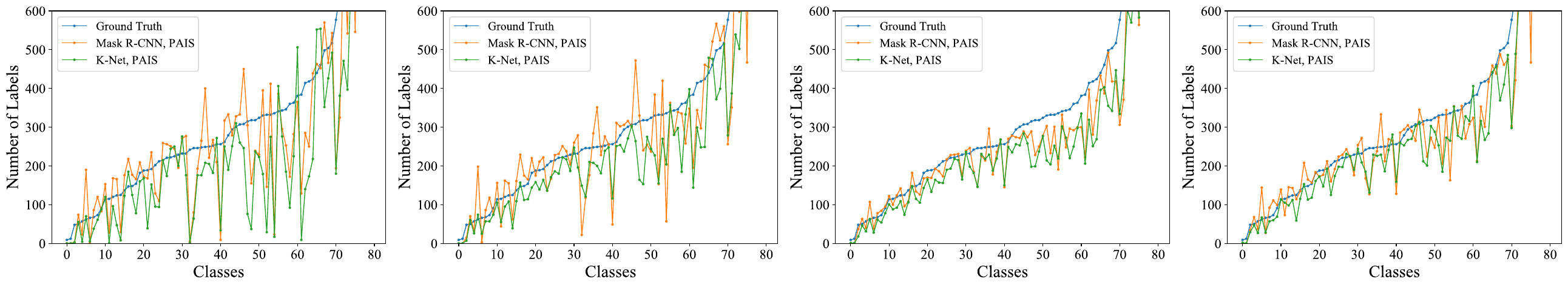}\vspace{-1mm}
\caption{
\textbf{Influence of imbalanced classes to semi-supervised training}.
The models are trained under 10\% labeled images for the results.
We randomly sampled 10k images from the rest of COCO \texttt{train2017} to draw these figures.
As the training progresses, the predictions from the teacher model are gradually fitted to the distribution of the number of labels.
}\label{fig6}
\end{figure*}
\textbf{Hyper-parameters in DALoss.} We investigated the effect of hyper-parameters, $\alpha$ and $\beta$, used in DALoss.
To give equal weight to both classification and segmentation, we set $\alpha=\beta$.
The results are shown in Tab.~\ref{tab6}, indicating that larger values of $\alpha$ and $\beta$ lead to better performance.
We also analyzed the functions of different hyper-parameters used to adjust the input score, as shown in Fig.~\ref{fig3}.
The figures illustrate that increasing the hyper-parameter enlarges the difference between high- and low-quality masks or classes, adjusting the input score more appropriately based on their quality.
However, when the hyper-parameter is set too large, the low-quality masks or classes are significantly constrained, potentially affecting the generalization of the model as noise is removed entirely during training.
\textbf{Model convergence speed.} In Fig.~\ref{fig4}, we analyze the convergence speed of the models.
The results demonstrate that DALoss can expedite the model training in terms of convergence rate.
For instance, Mask-RCNN (PAIS) achieves a convergence speed that is approximately 2 times faster than that of Mask-RCNN (PAIS w/o DALoss).
Moreover, by comparing the results between Mask-RCNN (PAIS) and K-Net (PAIS), we observe that DALoss may be particularly effective for box-dependent instance segmentation frameworks, which facilitates rapid convergence.
\textbf{Different thresholds for classification scores and mask IoUs.}
We observed from Tab.~\ref{tab7} and Tab.~\ref{tab8} that increasing the thresholds for classification score and mask IoU leads to a decrease in performance.
This suggests that DALoss is able to make use of misaligned pseudo-labels instead of simply filtering them out.
By allowing the model to learn from these potentially noisy labels, it is able to better handle situations where the alignment between labeled and unlabeled data is not perfect.
This may result in improved generalization to new, unseen data, as the model has learned to adapt to the presence of noisy labels.

\textbf{Different ratios of labeled and unlabeled images per batch.}
Tab.~\ref{tab9} shows that increasing the ratio of unlabeled images in a batch can improve the model's performance.
The results indicate that performance saturates when the ratio is set to 1:4, suggesting that adding too many unlabeled images may lead to diminishing returns.

\textbf{Visualizations.} We visualize the outputs of different models, including K-Net (supervised), K-Net (PAIS w/o DALoss), K-Net (PAIS), Mask-RCNN (supervised), Mask-RCNN (PAIS w/o DALoss), and Mask-RCNN (PAIS), in Fig.~\ref{fig5}.
The results show that: (1) PAIS can improve the recall for most instances in the supervised model, but the quality of the predictions is not guaranteed. (2) The use of DALoss in PAIS helps to improve the quality of predictions in terms of mask and classification.

\textbf{Influence of imbalanced classes.} In Fig.~\ref{fig6}, we investigate the effect of imbalanced classes on PAIS. We plot the number of labels obtained from the ground truth and the teacher models for Mask-RCNN and K-Net with PAIS at different iterations, namely 32k, 64k, 120k, and 180k. The results indicate that the predicted pseudo-labels gradually conform to the distribution of the imbalanced labels.

\section{Discussion}
\begin{table}[t]
\centering
\tablestyle{4.6pt}{1.13}\begin{tabular}{c|c|c|c|c|c}\shline
$\tau_{cls}$, $\tau_{iou}$ & 0.35, 0.3 & 0.50, 0.3 & 0.65, 0.3 & 0.35, 0.5 & 0.35, 0.7 \\ \hline
w/o DALoss &  29.3  & 29.0 & 25.3 & 29.3 & 29.0   \\
w/ DALoss & 31.1 & 30.6 & 29.9 & 30.8 & 30.6   \\ \shline
\end{tabular}\vspace{1.5mm}
\caption{DALoss \wrt thresholds tuning. DALoss consistently outperforms threshold tuning in all settings.\label{tab10}}
\end{table}

\begin{table}[t]
\centering
\tablestyle{3.0pt}{1.13}\begin{tabular}{c|c|c|c|c}\shline
Method & 1\% & 5\% & 10\% & 100\% \\ \hline
K-Net, supervise & 11.6 & 22.3 & 26.5 & 38.4   \\
EMA w/o DALoss & 17.8\textit{(+6.2)} & 25.4\textit{(+3.1)} & 29.3\textit{(+2.8)} & 39.4\textit{(+1.0)} \\
EMA w/ DALoss & 19.8\textit{(+8.2)} & 27.5\textit{(+5.2)} & 31.0\textit{(+4.5)} & 40.8\textit{(+2.4)}  \\
Performance gain & \textit{+2.0} & \textit{+2.1} & \textit{+1.7} & \textit{+1.4}  \\ \hline
M-RCNN, supervise & 11.5 & 22.4 & 27.1 & 37.5   \\
EMA w/o DALoss & 20.1\textit{(+8.6)} & 27.4\textit{(+5.0)} & 29.8\textit{(+2.7)} & 38.4\textit{(+0.9)}   \\
EMA w/ DALoss & 21.1\textit{(+9.6)} & 29.3\textit{(+6.9)} & 31.0\textit{(+3.9)} & 39.5\textit{(+2.0)} \\
Performance gain & \textit{+1.0} & \textit{+1.9} & \textit{+1.2} & \textit{+1.1}  \\ \shline
\end{tabular}\vspace{1.5mm}
\caption{Comparison of EMA with DALoss. The performance improvements are not mainly attributed to EMA.\label{tab11}}
\end{table}

\textbf{DALoss \wrt Thresholds Tuning.} As shown in Tab.~\ref{tab10}, we conduct an ablation study by removing DALoss from our method in Tabs.~\ref{tab7} and~\ref{tab8}.
The results show that DALoss consistently outperforms threshold tuning in all settings, which validates that DALoss is indeed more effective than tuning the thresholds.

\textbf{Comparison of EMA with DALoss.} We show that the performance improvements are not mainly attributed to EMA. First, we summarize the results of Tabs.~\ref{tab3} and~\ref{tab4} in Tab.~\ref{tab11}. We can find that DALoss achieves a larger performance gain than EMA in 100\% COCO (+1.4, +1.1 \vs +1.0, +0.9). Second, EMA alone cannot selectively utilize noisy pseudo-labels for better learning. DALoss solves this problem and leads to further improvement over EMA.

\textbf{Discussion on Score Filtering.} We have carefully chosen the values of $\tau_{cls}, \tau_{iou},$ based on Fig.~\ref{fig1}(c)(d), which illustrates that decreasing the thresholds will generate more noisy but informative pseudo-labels. Therefore, we use low thresholds to obtain such pseudo-labels, which is different from previous methods that need to adjust thresholds to filter them out. This simplifies the tuning of the thresholds. We believe that the experiments in Tabs.~\ref{tab7}, ~\ref{tab8}, and~\ref{tab10} are sufficient to show that lower thresholds are suitable for DALoss.

\textbf{Generality to Other Segmentation Tasks.} DALoss can be applied to other semi-supervised segmentation tasks, as the noisy pseudo-labels are common in semi-supervised setting. To show the generality, we apply DALoss to panoptic segmentation, which is a more challenging task that requires both instance and semantic segmentation. We report a preliminary result under 10\% COCO. The initial result shows that DALoss can improve the PQ performance from 36.8\% PAIS w/o DALoss to 37.3\% PAIS w/ DALoss.
We remain more experiments in our future work.

\textbf{Discussion on Large Segment Everything Models.} We envision that the future of image segmentation will not only aim to segment everything, but also to provide fine-grained text descriptions for the segmented regions. However, the recently proposed models such as SAM~\cite{kirillov2023segment} and SEEM~\cite{Zou_Yang_Zhang_Li_Li_Gao_Lee} either lack labeled semantic information or demand large amounts of labeled semantic data for training. Therefore, we believe that semi-supervised learning will be a crucial solution to leverage abundant unlabeled data and reduce the labeling burden.

\section{Conclusion}
In this paper, we presented a novel PAIS framework for semi-supervised instance segmentation.
To address the misalignment between classification score and mask quality, we introduced a dynamic aligning loss (DALoss), which aligns the classification loss term and the segmentation loss term based on the quality of different pseudo-labels.
Our experimental results demonstrate the effectiveness of the proposed PAIS framework.
Specifically, when the amount of labeled data is extremely limited, our pipeline equipped with PAIS and DALoss achieves superior performance for instance segmentation.
We believe that PAIS can serve as a strong baseline for future research on semi-supervised instance segmentation. We hope our work can inspire further exploration in this exciting research direction.

\vspace{-4mm}\paragraph{Acknowledgements.}
\footnotesize{This work was supported by National Key R\&D Program of China (2022ZD0118202), National Science Fund for Distinguished Young Scholars (No.62025603), National Natural Science Foundation of China (No.U21B2037, No.U22B2051, No.62176222, No.62176223, No.62176226, No.62072386, No.62072387, No.62072389, No.62002305 and No.62272401), and Natural Science Foundation of Fujian Province of China (No.2021J01002, No.2022J06001).}

{\small
\bibliographystyle{ieee_fullname}
\bibliography{egbib}
}

\end{document}